\documentclass[preprint]{IEEEtran}
\IEEEoverridecommandlockouts
\usepackage{cite}
\usepackage{amsmath,amssymb,amsfonts}
\usepackage{graphicx}
\usepackage{textcomp}
\usepackage{xcolor}

\usepackage{hyperref}
\usepackage[nolist]{acronym}
\usepackage{svg}
\usepackage{multirow}
\usepackage{subfig}
\usepackage{gensymb}
\usepackage{booktabs}

\usepackage{algpseudocode}
\usepackage[ruled]{algorithm}
\usepackage{algorithmicx}

\def\BibTeX{{\rm B\kern-.05em{\sc i\kern-.025em b}\kern-.08em
    T\kern-.1667em\lower.7ex\hbox{E}\kern-.125emX}}
\begin{document}

\title{Ternarization of Vision Language Models for use on edge devices}


\author{\IEEEauthorblockN{Ben Crulis,
Cyril De Runz, Barthelemy Serres\ and
Gilles Venturini}
\IEEEauthorblockA{LIFAT,
University of Tours\\
Blois, France}}

\maketitle

\begin{abstract}
We propose a process to compress a pre-trained Vision Language Model into a ternary version of itself instead of training a ternary model from scratch. A new initialization scheme from pre-trained weights based on the k-means algorithm is proposed to reduce the ternarization time.
We implement different custom operators for executing the ternary model on the TensorFlow Lite Engine. We compare the original model with its ternary and binary versions in terms of memory consumption, inference speed and perplexity.
We find that the ternary model using our custom ternary matrix multiplication operator provides a good compromise in term of memory usage and perplexity, while having the fastest token generation speed.
\end{abstract}

\begin{IEEEkeywords}
deep learning, ternary, edge device, quantization
\end{IEEEkeywords}

\begin{acronym}
\renewcommand{\\}{}
\acro{AI}{Artificial Intelligence}
\acro{ANN}{Artificial Neural Network}
\acro{API}{Application Programming Interface}
\acro{AVX}{Advanced Vector Extensions}
\acro{BNN}{Binary Neural Network}
\acro{BP}{Backpropagation}
\acro{CNN}{Convolutional Neural Network}
\acro{CPU}{Central Processing Unit}
\acro{CV}{Computer Vision}
\acro{DFA}{Direct Feedback Alignment}
\acro{DL}{Deep Learning}
\acro{DNN}{Deep Neural Network}
\acroplural{DNN}[DNNs]{Deep Neural Networks}
\acro{DRTP}{Direct Random Target Propagation}
\acro{FPGA}{Field-Programmable Gate Array}
\acro{GAN}{Generative Adversarial Network}
\acro{GD}{Gradient Descent}
\acro{GPS}{Global Positioning System}
\acro{GPU}{Graphical Processing Unit}
\acro{HCI}{Human-Computer Interface}
\acro{HTML}{HyperText Markup Language}
\acro{JVM}{Java Virtual Machine}
\acro{LCE}{Larq Compute Engine}
\acro{LLM}{Large Language Model}
\acro{ML}{Machine Learning}
\acro{MLP}{Multi Layer Perceptron}
\acro{PTQ}{Post Training Quantization}
\acro{QAT}{Quantization Aware Training}
\acro{RAM}{Random Access Memory}
\acro{RNN}{Recurrent Neural Network}
\acro{SGD}{Stochastic Gradient Descent}
\acro{SIMD}{Single Instruction, Multiple Data}
\acro{SOTA}{State Of The Art}
\acro{SOTA}{State Of The Art}
\acro{SSE}{Streaming SIMD Extensions}
\acro{SSL}{Self-Supervised Learning}
\acro{STE}{Straight-Through Estimator}
\acro{SfM}{Structure from Motion}
\acro{TDID}{Target-Driven Instance Detection}
\acro{TNN}{Ternary Neural Network}
\acro{VLM}{Vision Language Model}
\acro{VQA}{Visual Question Answering}
\acro{ViT}{Vision Transformer}
\acro{QAT}{Quantization-Aware Training}
\acro{PTA}{Post-Training Quantization}
\end{acronym}

\section{Introduction}

\acp{VLM} appear as the ideal type of \ac{DNN} to solve many vision tasks. Indeed a single \ac{VLM} can be used for tasks such as captioning, object detection, grounding or segmentation \cite{xiao2024florence}.

However, as they are built on top of \acp{LLM}, \acp{VLM} are often very large, with a number of parameters often ranging from the billions~\cite{liu2023llava,DBLP:journals/corr/abs-2407-07726} to the hundreds of billions~\cite{achiam2023gpt}.
The large size of these models makes them difficult if not impossible to use on edge devices such as smartphones. Even smaller \ac{VLM} models such as the Moondream model are too large to run on smartphones despite only having about $2$ billions parameters.

Fortunately, several methods have been developed in the domain of model compression to reduce the size of models while preserving their task performance. For instance, model distillation consists in training a smaller model to reproduce the outputs of a larger teacher model, effectively compressing the knowledge of the teacher model \cite{hinton2015distilling, polino2018model}. On the other hand, pruning removes the less useful weights of the model by setting their value to $0$, reducing the parameter count and thus the size \cite{lecun1989optimal}. Finally, quantization reduces the number of bits per parameter, effectively reducing the model size \cite{polino2018model}.
The most extreme form of quantization is the binarization of the weights\cite{Hubara2016}, which makes each parameter store only a single bit of information.

Unfortunately, as far as we know, no method for binarizing pre-trained models exists. This mean we have no choice but to train binary \acp{VLM} from scratch, which would still be long, difficult and expensive. A similar problem arises in model distillation, which requires the training of a smaller model from scratch.

We thus settle on a compromise: we propose to compress an already relatively small \acp{VLM} into a ternary version of itself in a process called \textit{ternarization}. In ternarization, the ternary weights of the compressed model are initialized from the original unquantized weights. The ternary weights allow the resulting model to retain slightly more information from the original model compared to binary weights, while still being extremely compressed and enabling potential for computational optimization. In ternarization, a large part of the original weights are set to $0$, which means ternarization can also be considered a form of pruning.

Overall, our proposed procedure consists in ternarizing and fine-tuning the original model in the PyTorch ecosystem. To initialize the parameters of the ternary model, we propose an algorithm that uses vector quantization to reduce the initial quantization error to reduce the amount of fine-tuning required. The resulting model is converted to the Tensorflow Lite format to make it compatible with edge device execution. This model is then compared to the original model as well as other variants in term of execution speed, memory usage and task performance. These variants differ by their use of different implementations of the ternary matrix multiplication algorithm.

We openly release the conversion code as well as optimized operators to unpack the ternary weights or directly compute the matrix multiplication with a packed ternary matrix.

\section{Related Work}

\ac{QAT} and \ac{PTQ} techniques are defined and surveyed in~\cite{gholami2022survey}. Our work inscribes itself in the \ac{QAT} paradigm as we re-train the model using calibration dataset, whereas \ac{PTQ} would completely avoid fine-tuning. Furthermore, $2$ bits quantization should not be confused with ternarization, as $2$ bits quantization make parameters use all possible $4$ values that can be encoded using $2$ bits. Such a quantization is also performed block-wise on the weight matrix, contrary to ternarization which transforms the entire weight matrix of a layer at once.

The ternarization of \acp{DNN} was explored before, but most research works train the ternary models form scratch instead of starting from pre-trained full-precision models. For instance, \cite{he2019simultaneously} uses a truncated gaussian approximation and \cite{yang2020harmonious} combine quantization and pruning techniques to train a ternary \ac{CNN} model (ResNet).

\cite{he2019optimize} Initializes the weights of their ternary model from the weights of a pre-trained model, but still requires a long training process over tens of epochs to recover the accuracy of the original model. In contrast, we stop our ternarization process after $2$ epochs of fine-tuning.
The authors of \cite{yang20241} ternarize a diffusion model for image generation. Similar to us, they implement custom low-level kernels to accelerate computations. However, they do not describe the ternarization process or release their code, which makes it impossible to compare their approach with ours in details.
Contrary to \cite{ma2024era} which also ternarize a \ac{LLM}, we ternarize an existing model from its pretrained weights using a novel initialization method instead of training the ternary \ac{LLM} from scratch.

\section{New Ternarization Process} \label{sec:ternarization}

The ternarization process starts from a pre-trained \ac{DNN} model.
The goal of the ternarization process is to assign most of the original weights of the model into one of the values of the set $\{-1,0,1\}$ with the smallest performance loss possible (measured in term of perplexity).

\subsection{Initialization of the ternary weights}

In order to extract the largest amount of information from the original model and thus accelerate the ternary fine-tuning process, a transfer learning approach like ours should initialize the latent weights to the values of the original model as it is possible in this case. However, doing this naively makes the initial model perform very poorly due to the high quantization error caused by the ternarization of the weights. Indeed, in general there is no reason to expect the original fine tuned weights to be uniformly distributed in the $3$ bins corresponding to the ternary values of the set $\{-1,0,1\}$ and have minimal quantization error from the start.

To reduce this problem and speed up the ternary fine-tuning, we resort to vector quantization. More specifically, we propose to use a modified version of the K-means algorithm~\cite{macqueen1967some,steinhaus1956division} to find the scaling parameter values that reduce the quantization error at initialization. The K-means algorithm is known to produce a solution that minimizes the quantization error, in fact, the objective function of the K-means algorithm to be minimized is precisely the mathematical definition of a quantization error in Euclidean space. Other vector quantization algorithms could be used in the initialization algorithm, for instance to avoid getting trapped in the potential local optima that the K-means optimization might fall in. Our complete initialization algorithm is described in Algorithm \ref{alg:kmeans_ternary}.

This algorithm can be interpreted as running a K-means algorithm in one dimension with the weights as the $1$D data points, with $k=3$ and two additional constraints. The first constraint is that one cluster centroid is always set to $0$ and the second constraint is that the two remaining cluster centroids have to be opposite of each other ($\mu_1 = -\mu_2$). The first constraint is required to avoid having to re-center the latent weights later and have yet additional parameters. The second constraint stems from the fact that we need equally spaced clusters because our staircase function $Tern$ of Equation \ref{eq:staircase} is odd and so the steps of the staircase function are themselves equally spaced. Having these two constraints make the algorithm simplify into the iterative computation of a single centroid corresponding to a free cluster. As a results, the algorithm only optimizes a single centroid parameter $\mu$. The algorithm minimizes the following modified K-means objective function:

\begin{multline}
    f(\mu) = \sum_{i=1}^n \mathbf{1}_{w_i > \frac{1}{2}\mu}(w_i - \mu)^2
     + \sum_{i}^n \mathbf{1}_{w_i < -\frac{1}{2}\mu}(w_i + \mu)^2 \\
     + \sum_{i}^n \mathbf{1}_{|w_i| < \frac{1}{2}\mu}w_i^2
\end{multline}

where $w_i$ is the $i$-th weight parameter of the neuron, $n$ is the number of parameters, $\mathbf{1}$ is the indicator function and $\mu$ the centroid parameter we optimize. $\mu$ will also be used to compute the initial scaling factor $s$ for the neuron. The first two terms correspond to the quantization error of the points in the two symmetric clusters and the last term to the quantization error of setting the remaining parameters to the fixed cluster centroid at $0$.

The algorithm first initialize the centroid of the free cluster to the mean of the absolute values of the neuron weights to provide a first approximation as can be seen at line \ref{alg:init_kmeans_abs_mean} of Algorithm \ref{alg:kmeans_ternary}. The purpose of the absolute value is to ensure that we find a symmetric solution for the clusters around $0$ by constraining two cluster centroids to be equal in the absolute value space.
Then, in a loop we iteratively compute the assignment of the weights to one of the three clusters. As we simplified the problem using constraints, we only need to compute the assignments of the single free parameter controlling the two opposite clusters at once. To do this, we compute the signed distance $e_j$ of each weight to the free cluster centroid, which can be interpreted as an error. Next, this signed distance is used to decide the assignment to the free cluster by checking if the weight is closest to $0$ or to the current centroid value. Then, the new free cluster centroid is computed as the mean of the weights belonging to the cluster.

Finally, the scaling factors and the scaled weights are returned. The scaled weights are used as latent weights that are ternarized at each forward pass and multiplied by the learned scaling parameter to get the weights used in the ternary matrix products. The learned scaling parameters are themselves initialized as the inverse of the scaling factor $s=\frac{1}{\mu}$ found by Algorithm \ref{alg:kmeans_ternary}. Since the central cluster is always centered are $0$, the threshold between the central cluster and one of the two outer clusters is always equal to half the absolute value of the outer cluster centroid value. In turn, since the neurons are scaled by the inverse of the centroid value which is the scaling factor $s$, this essentially makes $0.5$ the natural choice of threshold for the ternary staircase function of Equation \ref{eq:staircase}.

\begin{algorithm}
\algblockdefx[Input]{Input}{End}%
    {\textbf{Input:}}%
    {}

\begin{algorithmic}[1]
\Input{}
    \State $w$: the original weight vector
    \State $i$: the number of iterations of the algorithm
\End

\State $\mu \gets \frac{1}{n} \sum_j^n \vert w_j \vert$ \label{alg:init_kmeans_abs_mean}

\For{$t\in [1,\ldots,i]$}
    \State $\forall j, e_j \gets \vert w_j \vert - \mu$
    \State $\mu = \frac{\sum_j^n \vert w_j \vert \times \textbf{1}_{e_j > -\frac{1}{2}\mu}}{\sum_j^n \textbf{1}_{e_j > -\frac{1}{2}\mu}} $
\EndFor

\State $s \gets \frac{1}{\mu}$

\State \Return $s, w \times s$ \Comment{Return the scaling factor and the scaled latent weights}

\end{algorithmic}
\caption{Modified K-means algorithm for parameter initialization} \label{alg:kmeans_ternary}
\end{algorithm}

Algorithm \ref{alg:kmeans_ternary} only needs to be executed independently once for each neuron of the model at initialization. Clearly, as the number of clusters is constant ($3$), as the dimension of each point is $1$ and as each K-means step requires an amount of computation proportional to the number of weights, the time complexity of this algorithm is $O(ni)$ where $n$ is the number of weights and $i$ is the number of iterations required for convergence. This is simply the time complexity of K-means with $d$ set to $1$ and $k$ set to $3$.
In practice, we take advantage of the shared number of iterations to compute the initialization for all neurons of a layer in parallel thanks to a vectorized implementation of the algorithm.
Then the fine-tuning process takes over to further adjust the latent weights in a way that takes into account the inter-dependencies between neurons and the dependencies with the input data.
The algorithm's main loop theoretically need to run until convergence, but in practice we find that the algorithm converges in only $10$ iterations, as can be seen in Figure \ref{Fig:ternary_init}. We use $i=10$ iterations in the remainder of this chapter.

\subsection{Ternary weights training}

Once the ternary weights have been initialized from the original weights, the ternary weights are then fine-tuned using \ac{SGD}. Similar to binarization, the ternary weights are obtained from real valued \textit{latent} weights which are quantized to ternary values at each forward pass using a ternarization function which plays the same role as the $sign$ function in \acp{BNN}\cite{Hubara2016}. In this work we chose to use a simple staircase function producing ternary values defined as:

\begin{equation} \label{eq:staircase}
    Tern(x) =
    \begin{cases}
      -1, & \text{if}\ x < -0.5 \\
      0,  & \text{if}\ -0.5 \leq x < 0.5 \\
      1,  & \text{if}\ x \ge 0.5
    \end{cases}
\end{equation}

Just like the $sign$ function, the $Tern$ function is a staircase function that has a gradient that is $0$ almost everywhere, so training requires the use of a \acf{STE} as in \cite{Hubara2016}. In this work, we choose the identity function as the \ac{STE} of this staircase function. Furthermore, we constrain the latent weights into the $[-1, 1]$ interval (clipping) after each update.

The ternarization of pre-trained weights can induce the apparition of infinite values in the activations of layers containing ternary weights. To remedy this problem, we introduce a learned scaling parameter vector and we clip the values of the activations as well. Each scaling parameter controls the scaling of all of the weights of a single neuron. The weight vector used for the ternary matrix multiplication is thus:

\begin{equation}
    w^* = s \times Tern(w)
\end{equation}

Where $w$ is the \textit{latent weight} vector and $w^*$ is the final weight vector used to compute the activation of the neuron.
In practice, at inference we use the linearity of the matrix product to delay the scaling operation to let us use a fast ternary matrix multiplication implementation.

Since the scaling factor is fine tuned, learning the scaling factor in the fine tuning step is equivalent to learning a per-neuron threshold for the ternary function $Tern(x)$.

\section{Experiments}

In our experiments, we use the Moondream$2$ model\footnote{\url{https://github.com/vikhyat/moondream}} released with its trained weight under a Apache-$2.0$ license. Moondream$2$ is a $1.6$ billion parameters model based on the Phi \ac{LLM} model family from Microsoft~\cite{abdin2024phi}, which is a family of \ac{LLM} oriented for use on edge device. The models handles images of size $378 \times 378$ which allows it to detect some details in the images.

The Moondream$2$ model is a \ac{VLM} composed of a \ac{ViT} image encoder and of a \ac{LLM} Transformer. The image encoder is used to compute an image embedding that becomes part of the prompt of the \ac{LLM} and allows it to use the visual information to form the generated response. More specifically, the input image is resized to a size of $224 \times 224$ and broken down into a sequence of $8 \times 8$ image patches. The patch sequence is processed by a transformer in a way that each patch is processed contextually to all other patches. For each image patch an embedding will be obtained in the output of the image encoder. The embeddings are designed to be of the same dimension as the token embeddings of the \ac{LLM}. These $784$ image embeddings are inserted in the prompt of the \ac{LLM} along with the other token embeddings of the prompt containing for instance the user question among other information. 

In this work, we focus on ternarizing only the \ac{LLM} part of the model as the image encoding part is treated as a pre-processing step that has negligible inference cost in comparison to the \ac{LLM} token processing and generation. In principle, nothing prevents us to apply the same ternarization procedure to the image encoder module.

We choose to use Moondream$2$ because it is one of the smallest \acp{VLM} in term of number of parameters, which makes it a great candidate for use on edge device. However, without any type of quantization, this model would still be too large to run on low to medium end smartphones. After ternarization, it is possible to run the model on low-end smartphones with about $1$GB of free memory.

\subsection{Ternarization}

We set ternary weights for all of the dense linear layers of each Transformer block of the model, except the first and the last. As can be seen in Figure \ref{fig:quantization_details}, the blocks $2$ to $23$ are the only blocks to be ternarized. The first and last blocks are left untouched to avoid introducing quantization error too early in the embeddings and allow the model to decode the embeddings more easily in the last few layers. Indeed, recent research suggests that early and latter layers perform critical transformations and cannot be removed or perturbed without significantly degrading performance~\cite{lad2024remarkable}.
The token embedding and classification layers are also not ternarized. However, we still perform int$8$ quantization on the remaining parameters to produce the final models, which are found in the transformers block $1$ and $24$ as well as in the embedding and classifier layers as can be seen in Figure \ref{fig:quantization_details}.
This ternarization step adds a negligible amount of scaling parameters that will be fine tuned alongside the original parameters. For the fine-tuning, we set all weights as trainable, including the original weights that have been left untouched, so they can adapt to the other ternarized weights. We do not fine tune the parameters of the vision encoder at all, only the parameters of the LLM transformer part of the \ac{VLM}.

\subsection{Protocol}

We now describe the protocol for our experiments comparing the base model with binarized and ternarized version of the original model.

We fine tune the ternary model for $2$ epochs on the \textit{conversation\_58k} subset of the LLaVA dataset~\cite{liu2023llava} (LLaVA Visual Instruct $150$K Dataset\footnote{\url{https://huggingface.co/datasets/liuhaotian/LLaVA-Instruct-150K}}) that we treat as our calibration dataset or train set. The LLaVA dataset is comprised of thousands of synthetic conversations between two agents: one agent denoted as ``human'' and another agent denoted as ``AI assistant''. The conversations of the dataset reference images from the COCO dataset~\cite{lin2014microsoft}. Most of the conversation will have the human agent ask questions about the images and have the AI assistant answer based on the visual feedback from the referenced image associated with the conversation. The LLaVA dataset is a synthetic dataset as the data was collected by prompting OpenAI's GPT-$4$ model. For training and evaluating, the loss is computed specifically on the tokens predicted by the assistant excluding the tokens correspond to the human messages, evaluating its ability to predict the tokens of what is considered a gold standard response in our experiments.

The goal of this fine-tuning step is not to give the model new capabilities but instead to make it recover most of the test performance of the original model that was lost when applying the staircase $Tern$ function on the weights. We find that training for $2$ epochs is sufficient to reach convergence in our experiments. For this step, we use a batch size of $8$ to fit the memory of our graphics card with $2$ steps of gradient accumulation. We use a base learning rate of $1e^{-3}$ and use a cosine annealing schedule to match the recommended fine-tuning settings for this model. As we fine tune a ternary version of the model and not the original model itself, a full hyperparameter optimization step should probably be performed again to ensure the most efficient hyperparameters are used but we skip it to save on the amount of computations as it would imply training tens to hundreds of models.

Along with the ternary model, we also train a binary model to compare the performance in term of perplexity and file size. The binary model will serve as a reference lower bound for the most extreme type of quantization we can apply to the model. As the binary model ends up losing too much performance in term of perplexity, we do not consider it for the comparisons in term of memory usage and inference speed.

\begin{figure}[t]
    \centering
    \includegraphics[width=\linewidth]{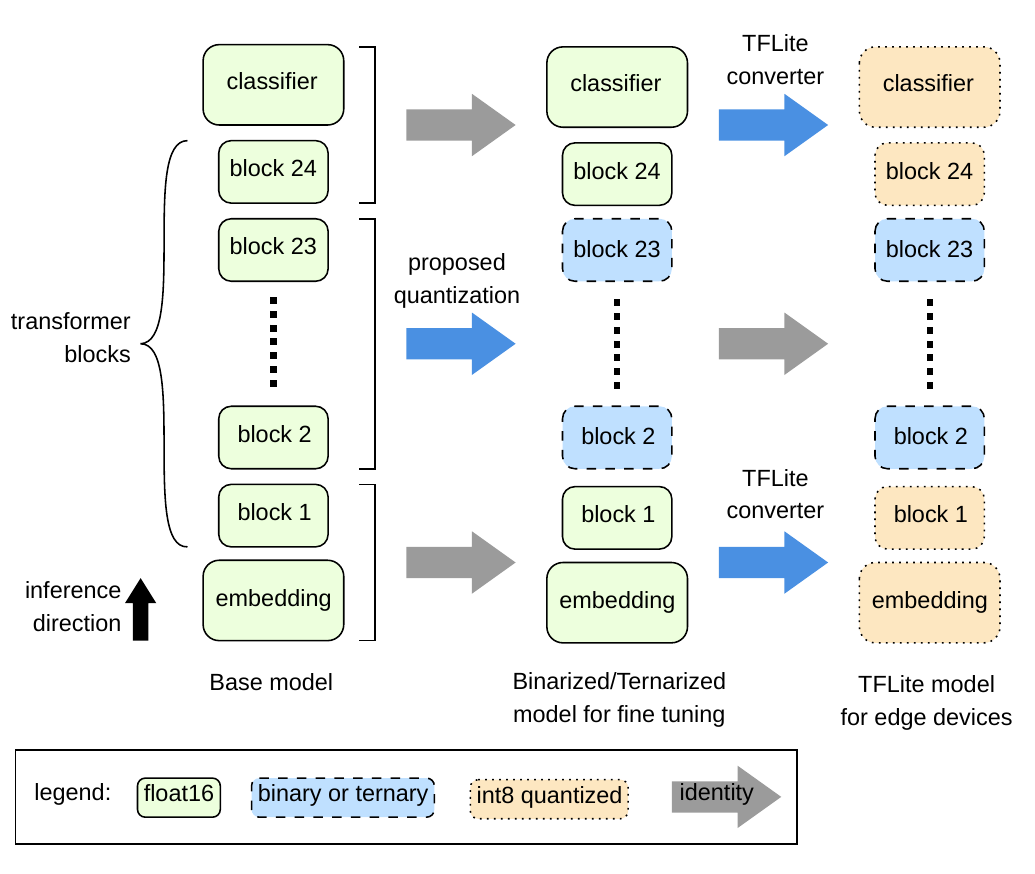}
    \caption{Details of the quantization process of the \ac{LLM} Transformer.}
    \label{fig:quantization_details}
\end{figure}

We convert the fine tuned ternary model from PyTorch to the Tensorflow format. Finally, we export the ternarized Tensorflow model to Tensorflow Lite, which is the format used for high performance \ac{DNN} inference on both desktop and edge devices through the TFLite Inference Engine. Tensorflow Lite is the only mature high performance neural network inference engine allowing deployment of \acp{DNN} models on smarpthones. Alternatives include the deprecated \textit{PyTorch Mobile}\footnote{PyTorch Mobile: \url{https://pytorch.org/mobile/home/}} and the recently released \textit{ExecuTorch}\footnote{ExecuTorch: \url{https://pytorch.org/executorch-overview}} at the time of writing.
Before exporting to TFLite, we add a final int$8$ quantization step to compress the remaining float$32$ weights of the linear layers that have not been ternarized. The details of the quantization are illustrated in Figure \ref{fig:quantization_details}.

We evaluate the model on the \textit{detail\_23k} subset of the LLaVA dataset. The original PyTorch model and the TFLite models are evaluated on \ac{CPU} as it is most representative of the capabilities of low and medium end smartphones.
To avoid long computational times due to CPU usage, we sample $100$ conversations from the \textit{detail\_23k} subset and compute the mean and standard deviation.

The ternarized model is further split into $3$ variants: q2-tf, q2-unpack and q2-matmul.
q2-tf is the ternarized model which uses only readily available TFLite operators to unpack the ternary weights before computing the matrix multiplications with the native TFLite batch matrix multiplication operator. The q2-unpack model is the same except the unpacking operation is carried out by a custom operator with a native implementation in C++.
Finally, q2-matmul uses another custom operator with a native multi-threaded implementation with additional optimizations to perform the batch matrix multiplication using the ternary weights directly.

This split will allow us to know if implementing optimized native operators for ternary models provides advantages in term of speed and memory consumption.

Overall, we summarize the different model variants using the following codenames:

\begin{enumerate}
    \item \textbf{base}: the original pre-trained PyTorch model
    \item \textbf{binary}: the fine tuned binary model in PyTorch
    \item \textbf{ternary}: the fine tuned ternary model in PyTorch
    \item \textbf{tf~(qint8)}: the fine tuned original pre-trained model converted to the TFLite format with int8 quantization
    \item \textbf{q2-tf}: the fine tuned ternary model TFLite model that uses only TFLite or Tensorflow Operators
    \item \textbf{q2-unpack}: the fine tuned ternary TFLite model that uses a custom \textit{unpack} operator
    \item \textbf{q2-matmul}: the fine tuned ternary TFLite model that uses a custom \textit{ternary matrix multiplication} operator
\end{enumerate}

q2-tf, q2-unpack and q2-matmul all have quantization schemes corresponding to the illustration of the TFLite model of Figure \ref{fig:quantization_details}, except tf~(qint8) which has all of its transformer blocks and layers quantized in the int8 format.

\subsection{Implementation details}

We implement our custom operators in a fork of the Larq Compute Engine~\cite{larq} which is built on top of Tensorflow Lite. The Larq Compute Engine (LCE) already implements some highly optimized operators for \acp{BNN}, but no custom operator for Ternary Neural Networks. However, this ecosystem allows for the easy addition of new custom operators for Ternary layers.

The custom operators use \ac{SIMD} instructions, which are processor-specific instructions that allow to perform vectorized operations such as adding multiple values in parallel on \ac{CPU} very efficiently. More specifically, on desktop \acp{CPU}, we leverage the \ac{SSE} and \ac{AVX} instructions sets.

\section{Results}

We know present the results we obtain in term of perplexity, memory usage and execution speed of the resulting ternary model. The different metrics are measured on the \textit{detail\_23k} subset of the LLaVA dataset, which we use as the test set.

\subsection{Initialization method}

\begin{figure}[t]
\centering
\includegraphics[width=1.0\linewidth]{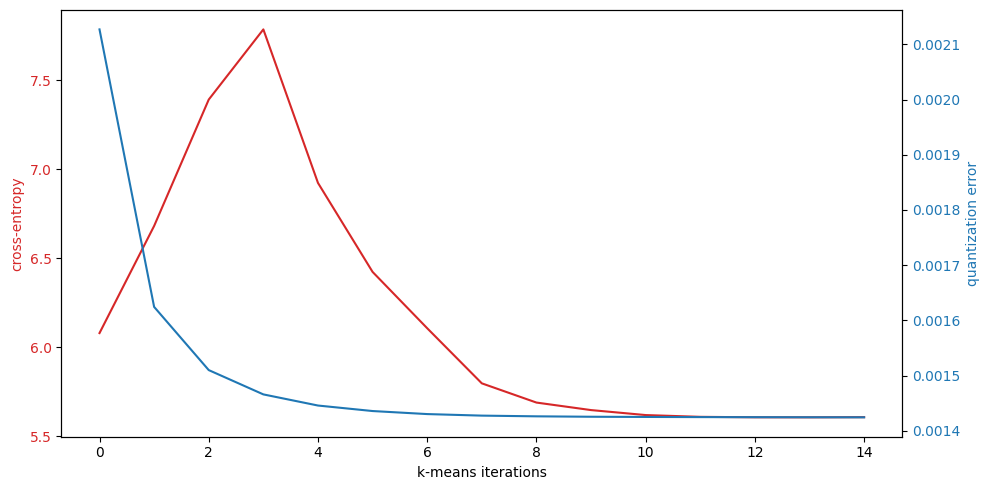}
\caption{Initial loss and quantization error by iteration of Algorithm \ref{alg:kmeans_ternary}}\label{Fig:ternary_init}
\end{figure}

We first verify that our proposed initialization scheme from pre-trained weights of Algorithm \ref{alg:kmeans_ternary} helps reducing the quantization error and the initial loss of the ternarized model.
In Figure \ref{Fig:ternary_init} we plot the cross-entropy loss on the test set as well as the average quantization error of the scaled ternarized weights. The average quantization error is computed as:

\begin{equation}
    err = \frac{1}{d} \sum_i^d (Tern(w_i)\times scale - w_i)^2
\end{equation}

where $w$ is the weight vector, $d$ is the number of parameters and $scale$ is the scale factor returned by the initialization algorithm.

As expected of a K-means based algorithm, the average quantization error decreases with more iterations of the algorithm. Yet, the quantization error does not reach $0$, but instead reaches a plateau after $7$ iterations. Since the quantization error does not reach $0$, we can expect the loss of the ternarized model to be higher than the base model after initialization.

Surprisingly, the loss first \textit{increases} with the number of iterations, before decreasing again and stabilizing to a lower loss than with $0$ iterations. We do not investigate this behavior further in these experiments. As the cross-entropy loss eventually reaches a lower value than initially, this initialization procedure essentially gives the fine-tuning process a head start at very negligible computational cost and without using any training data.

Overall, since both the quantization error and the loss reach a plateau after about $10$ iterations, we set the number of iterations to $10$ for the ternary models.

To decrease the loss further and recover a better text generation performance, we fine tune the model on a calibration dataset as explained in the next section.

\subsection{Fine Tuning}

The ternary model is initialized using $10$ steps of K-means following the initialization method described previously. The binary model does not require a special initialization. The ternary and binary models are then fine tuned on the calibration dataset (\textit{conversation\_58k}).

The binary model shows some signs of instabilities, the train and test loss suddenly increase around $2300$ steps. These kinds of instabilities also occurred a few times while developing the early development versions of this experiment, but did not happen to the ternary model.

Even after fine-tuning until convergence, the loss never quite reaches the same test loss as the reference unquantized model. This is consistent with previous work that also observe a loss of performance on \ac{CNN} and \ac{RNN} models in extreme quantization regimes even after the quantized models are fine tuned~\cite{boo2019memorization}. The given explanation is that quantization reduces the total \textit{capacity} of the model to remember information. In turn, a loss in performance due to quantization is caused by the original model having been trained to saturation of its capacity.

\subsection{Trade-offs between perplexity and memory costs}

We verify that the ternarization procedure does not dramatically increase the test perplexity and that the custom operator implementation is still providing the correct results.

\begin{figure}[t]
\centering
\includegraphics[width=1.0\linewidth]{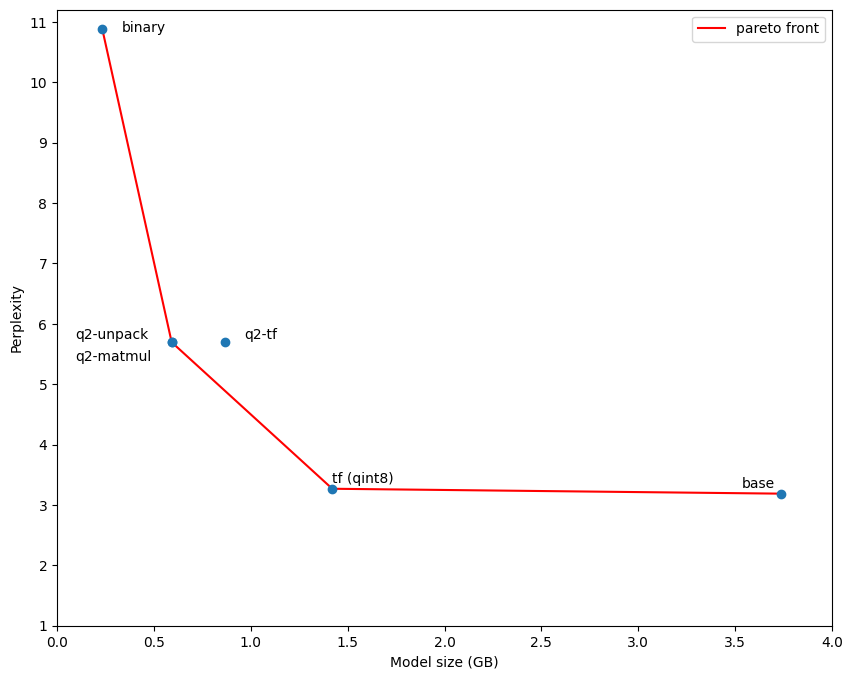}
\caption{Perplexity by file size (GB)}\label{Fig:perplexity_by_file_size}
\end{figure}

We examine perplexity as a function of the model file size on disk in Figure \ref{Fig:perplexity_by_file_size}.
The file size of the binary model variant is theoretical as we did not implement a Tensorflow version of the binary model with $1$bit packed weights.
A clear Pareto front appears with the binary model having the worst perplexity but the smallest file size at one end and the base model at another, with the smallest perplexity but ends up with the largest file size ($3.5$GB). The original PyTorch models all have parameters stored in float$16$ values.
Our ternary models with custom operations sit somewhere in the middle of the Pareto front, with a file size at $565$MB. Note that both points are overlapped in the figure.
The int$8$ model quantized using the TFLite converter more than halves the size of the original model ($1.4$GB) at a very negligible increase in perplexity, which makes another size-perplexity trade-off, even though in this case the increase in perplexity is barely measurable.

Finally, the q2-tf model appears to be slightly larger than the ternary models using custom operators, which puts it at $829$MB. We explain this increase in size by the addition of native tensorflow operators to dequantize the weights in the computational graph of the model, which makes optimizating the size of the model harder.
This model is the only variant that does not reach the Pareto front and is dominated by the custom operator variants.

\begin{figure}[t]
\centering
\includegraphics[width=1.0\linewidth,clip,trim={0 0 0 0.7cm}]{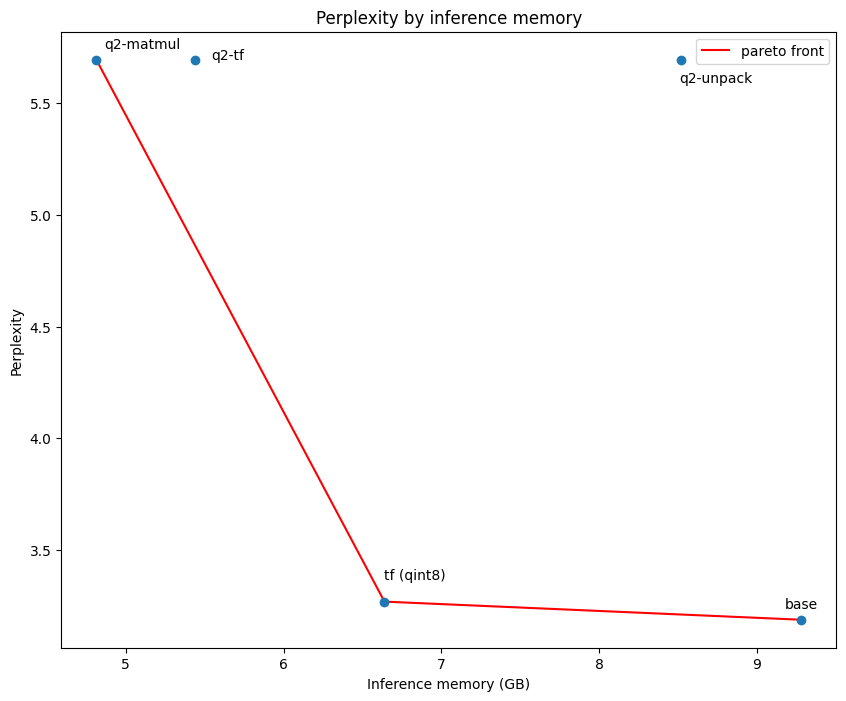}
\caption{Perplexity by model memory usage (GB)}\label{Fig:perplexity_by_inference_mem}
\end{figure}

We then report the perplexity as a function of the memory cost at inference in Figure \ref{Fig:perplexity_by_inference_mem}. The memory measured includes the model weights themselves, as well as the code require to run the models and the temporary tensors storing the intermediate of the computations. Due to its poor performance in term of perplexity, we did not make a Tensorflow implementation of the binary model and thus we do not include it either here or in the remaining tests.

This time, q2-unpack does not belong to the Pareto front due to its very high memory use at inference time. However, q2-matmul is the least costly of all variant tested at a memory usage of just under $5$GB on average. The base variant uses the most memory, using more than $9.2$GB on average to perform the computations.

The high cost of the q2-unpack variant is explained by the fact that, to perform the matrix multiplication, the custom operator first decompresses all of the weights of the ternary linear layers into float$32$ tensors, which uses a lot of memory. As the memory usage strategy is up to the TFLite engine, the memory space of the unpacked weights is not necessarily re-used between layers. As a consequence, the model takes almost as much memory as the original model. 

\subsection{Token generation benchmark}

Token generation is the procedure by which \acp{LLM} generate their responses. In the token generation phase, a \ac{LLM} first takes a prompt as input and then predicts a probability distribution on the next token. After this, a token is sampled from the distribution according to some specified procedure and the process starts over with the new token added to the context. The procedure repeats until a ``end of sequence'' token is predicted or when the maximum number of tokens is reached.

We measure the memory cost and the token generation speed while generating exactly $50$ tokens using the TFLite models. For the benchmark, we limit the number of CPU threads of the TFLite engine to $4$ for all models. The TFLite GPU delegate is completely disabled for these experiments. We select the most likely token as the predicted token at each step.
We report the results of the token generation benchmark in Table~\ref{tab:ternary_token_gen}.

\begin{table*}
\centering
\caption{Memory cost and token generation speed in TFLite} \label{tab:ternary_token_gen}
\begin{tabular}{rrrrrrrrr} 
\toprule
model variant        & \multicolumn{2}{c}{tf (qint8)}                     & \multicolumn{2}{c}{q2-tf}                          & \multicolumn{2}{c}{q2-unpack}                      & \multicolumn{2}{c}{q2-matmul}                       \\
                  & \multicolumn{1}{c}{mean} & \multicolumn{1}{c}{std} & \multicolumn{1}{c}{mean} & \multicolumn{1}{c}{std} & \multicolumn{1}{c}{mean} & \multicolumn{1}{c}{std} & \multicolumn{1}{c}{mean} & \multicolumn{1}{c}{std}  \\ 
\midrule
inference memory  & $1.64$                   & $0.15$                  & $1.25$                   & $0.11$                  & $5.18$                   & $0.48$                  & $\textbf{0.78}$          & $0.06$                   \\
tokens per second & $3.59$                   & $0.03$                  & $0.18$                   & $0.01$                  & $1.63$                   & $0.12$                  & $\textbf{8.78}$          & $0.17$                   \\
\bottomrule
\end{tabular}
\end{table*}

We observe that our proposed ternary model q2-matmul using the custom ternary multiplication operator provides the highest amount of token generated per second as well as the smallest memory footprint. More specifically, this variant uses $2$ times less memory as the second fastest model in term of token generation speed, which is the tf~(qint8) model which is only $8$bits quantized. This ternary variant is also more than two times faster at generating tokens compared to the tf~(qint8) variant.

The q2-unpack variant uses the most memory, more than $6$ times the amount of memory used by the best model, but places third on the token generation speed. 
Lastly, the q2-tf variant which only uses native Tensorflow operators to provide the weight unpacking manages to use less memory than the quantized base model tf~(qint8), but shows a very slow token generation speed at $0.18$t/s which is almost $50$ time slower than the fastest model.

This demonstrates how the difference of implementation of the ternary operators can largely affect the practical usefulness of ternary models. Our custom ternary matrix multiplication implementation enables the q2-matmul variant to be both the fastest and most memory efficient model even compared to the int$8$ quantized based model.

\section{Discussion}

Overall, q2-matmul is the most promising variant when the limiting factor is memory usage.
Excluding the binary model, this variant also corresponds to the file of the smallest size on disk, making it easily included and loaded in mobile applications.
The quantized tf variant which might seem a very good compromise at first still uses a large amount of memory which makes it almost unusable on most smartphones.

The different implementation of the ternary matrix multiplication or of the ternary weight unpacking algorithm does not change the perplexity, as expected of a correct implementation.

\paragraph{On the use of the TFLite Engine.}
The use of Tensorflow Lite as the inference engine seems inadequate for the kind of models used in our experiments. Indeed, firstly the TFLite does not natively support instructions for binary or ternary operations. Secondly, the TFlite engine is expected to be used for the inference of small models with a number of parameters in the hundreds of millions at most. We are using the engine to run the inference for a model with more than $1.6$ billions parameters, so we are pushing the limits of the abilities of the engine. Lastly, the TFLite engine primarily specializes in the execution of \acp{CNN}, not in the execution of Transformers which are a more recent architecture. In particular, of the Attention layers which benefit greatly from support of dynamic shape tensors of high rank and efficient cache management, which TFLite does not provide.
Nevertheless, the TFLite is the only mature engine compatible with mobile devices that can handle a novel use case such as the execution of \acp{VLM}. 

\paragraph{Ternarization versus Quantization.}
It is worth reminding that ternarization is not equivalent to quantizing the weights to $2$ bits for two main reasons. First, $2$ bit quantization make use of the $4$ possible values that can be encoded using two bits, whereas ternarization uses $2$ bits concretely, but only uses $3$ of the $4$ values, which means ternary parameters theoretically only take up to $log_2(3)\approx 1.585$ bits for storage. This means that ternary network weights can be compressed further if several weights are compressed as the same time using a general lossless compression algorithm.
Secondly, general quantization processes usually quantize matrices \textit{block by block}, which means that different scaling factors and zero points are used throughout the parameters of a given layer, whereas ternarization uses at most a single factor for each neuron and does not use a zero point. Both of these differences make ternary networks potentially faster to compute since the scaling operation can be delayed until it is applied to the result vector instead of being applied to the weight matrix which imply a much greater number of floating point number multiplications. Due to this difference in encoding scheme, Ternary network also possess the advantage of being computable using only addition and subtraction operations, which can be exploited in current and future hardware.

\paragraph{Even higher performance with ternary models.}
The ternary matrix multiplication also enables potential for software optimizations.
We mainly focus on \ac{CPU} execution since most consumer smartphones still do not have acceleration units that are compatible with \ac{DNN} computations. Unfortunately, the architecture of \acp{CPU} is far from ideal to perform dense neural network operations that are inherently parallel such as matrix multiplications, which are at the very heart of modern \acp{DNN}. This limitation is partially solved by the multi-core nature of modern \acp{CPU}, as well as by the addition of vectorized \ac{SIMD} instructions, which can process multiple values in parallel on a single core using a single \ac{CPU} cycle.

It is possible that even better performance could be obtained by making a better use of existing \ac{SIMD} instructions to reduce total latency and increase throughput when performing the dot products between floating point numbers and ternary values. In the future, the \ac{SIMD} instruction set might also grow to include more fused multiply-and-add operations, perhaps allowing to carry out the ternary dot product in as few as a single \ac{CPU} instruction.
Theoretically, the ternary matrix multiplication algorithm only involves adding or subtracting values from accumulators. This means that ternary models could also be implemented without requiring to load the ternary values from a weight tensor, but instead could be entirely implemented directly as machine instructions. In this case, the weights having a value of $0$ would simply not appear in the binary computer program, which would reduce the program size and the inference costs even further.

Additional performance improvements could be obtained on \acp{GPU} or \ac{GPU}-like processor architectures, at the condition that efficient instructions are available there as well to perform the ternary dot product.

The best possible performance would theoretically be obtained on specialized hardware or \acp{FPGA} that would compute the ternary dot product as combination of efficient sum or subtraction operations physically, which is something that can only be achieved with binary or ternary values.

\section{Conclusion}

We ternarized a pre-trained \ac{VLM} to compress its size and accelerate computations.

We achieve this by using an initialization procedure that re-use the original model weights to transfer knowledge from the original model. This initialization procedure minimizes the initial quantization error to enable faster training in the fine-tuning phase of the ternarization process. The fine-tuning process that follows uses a \ac{STE} to train the weights despite the quantization of most parameters and reaches acceptable performance in term of perplexity on the test set.

We show how the final ternary model can be run on the TFLite Interpreter, which is a platform to perform neural network inference on edge devices. We show how implementing custom native ternary operators can be beneficial in term of speed and memory consumption. More specifically, we show that in a token generation benchmark, the custom implementation using the ternary matrix product algorithm enables the ternary model to be twice as fast and twice as memory efficient as the second TFLite model, which is the original non-ternary model quantized by the TFLite converter.

Our ternarized \ac{VLM} model can be loaded locally on edge devices having a low as $1$ GB of free memory, despite the model still having more than $1$ billion parameters. The upper bound for the inference memory cost is higher, at around $3$GB, but this is still a reasonable amount of free memory requirement on modern middle-end to high-end smartphones.
However, the compression of the model to ternary weights comes at the cost of an increase in perplexity, which brings it to $5.5$ instead of the original $3.2$.

\bibliographystyle{IEEEtran}
\bibliography{references}

\end{document}